\documentclass[times,twocolumn,final,authoryear]{elsarticle}

\usepackage{framed,multirow}

\usepackage{amssymb}
\usepackage{latexsym}

\usepackage{url}
\usepackage{xcolor}
\definecolor{newcolor}{rgb}{.8,.349,.1}
\journal{Pattern Recognition Letters}

\begin{document}
\begin{frontmatter}
\title{Deep Learning Approaches for Image Retrieval and Pattern Spotting in Ancient Documents}

\author[1]{Kelly Lais Wiggers}
\ead{kelly.wiggers@pucpr.edu.br}
\author[1,2]{Alceu de Souza Britto Junior}
\author[3]{Alessandro Lameiras Koerich}
\author[4]{Laurent Heutte}
\author[5]{Luiz Eduardo Soares de Oliveira}

\address[1]{Pontifical Catholic University of Parana (PUCPR), Imac. Concei\c{c}\~{a}o, 1155 - Prado Velho, Curitiba (PR), Brazil}
\address[2]{State University of Ponta Grossa (UEPG), Av. Gen. Carlos Cavalcanti, 4748, Ponta Grossa (PR), Brazil}
\address[3]{\'Ecole de Technologie Sup\'erieure (\'ETS), Universit\'e du Qu\'ebec, 1100 rue Notre-Dame Ouest, H3C 1K3, Montr\'{e}al (QC), Canada}
\address[4]{Normandie Univ, UNIROUEN, UNIHAVRE, INSA Rouen, LITIS, Technopole du Madrillet, 76800 Saint-\'Etienne-du-Rouvray, France}
\address[5]{Federal University of Paran\'a, Av. Cel. Francisco H. dos Santos, 100 - Jardim das Américas, Curitiba (PR), Brazil}


\begin{abstract}
This paper describes 
two approaches for content-based image retrieval and pattern spotting in document images using deep learning. The first approach uses a pre-trained CNN model to cope with the lack of training data, which is fine-tuned to achieve a compact yet discriminant representation of queries and image candidates. The second approach uses a Siamese Convolution Neural Network trained on a previously prepared subset of image pairs from the ImageNet dataset to provide the similarity-based feature maps. In both methods, the learned representation scheme considers feature maps of different sizes which are evaluated in terms of retrieval performance. A robust experimental protocol using two public datasets (Tobacoo-800 and DocExplore) has shown that the proposed methods compare favorably against state-of-the-art document image retrieval and pattern spotting methods.  
\end{abstract}


\end{frontmatter}

\section{Introduction}
Content-Based Image Retrieval (CBIR) and Pattern Spotting (PS) are very active research topics in the field of Pattern Recognition. The main motivation is the increasing demand for solutions capable of performing the retrieval of an image, or specific objects in it, from wide digital libraries stored in the last decades of modern society. Some efforts in this direction are the recent studies described in \citep{DaoThiThuy201730,Wu2017,Xu201745,en}, which consist of finding relevant image candidates in digital collections based on a given query represented by the whole image or just by a pattern available on it.

A recent and exciting challenge on CBIR and PS has been the need for performing the retrieval task without any previous knowledge about the images or patterns to be retrieved. The idea is to produce generic solutions able to work on different digital image libraries. This is an interesting challenge that increases considerably the level of difficulty of the retrieval task. Thus, a robust solution must not only absorb the common image variability in terms of color, shape, conservation, quality, and context but must also consider the nonexistence of previous knowledge about the possible queries of a given image library.

The success of building such flexible solutions relies on: a) the definition of a robust representation for the image candidates and the queries, and b) the definition of an adequate distance metric able to estimate the similarity between a given query and the available image candidates. With respect to the representation, deep models, especially Convolutional Neural Networks (CNNs), are an attractive alternative to learn automatically a robust representation \citep{lecun2010}, avoiding the hard job of engineering related to the definition of handcrafted features. In fact, the use of deep models as a representation learner is currently a tendency. Several CNN architectures have been proposed in the last years, with applications on images, sounds, and texts. In particular for image retrieval, some significant contributions in terms of automatic representation using deep models are described in \citep{Babenko_2015_ICCV,joe,Gordo2016}. More than learning the representation, some works have shown the contribution to the image retrieval by using the concept of transfer learning, when a model trained on a large dataset such as the ImageNet is used, like in \cite{Babenko_2015_ICCV} and \cite{Fuzhen}. 

Regarding the similarity estimation between images, recent deep metrics have been used in different applications. Different from traditional architectures, a deep metric is usually composed of two deep models (CNN based) organized in a Siamese architecture. Such a deep architecture has shown successful results in different applications such as face verification \citep{Koch2015SiameseNN,8302003} and gesture recognition to predict if an input pair of images are similar or not. 

In this paper, we describe the two approaches based on deep learning that we have developed during the trajectory of our investigations on how to reduce the dependency of CBIR and PS solutions on previous knowledge of the image dataset in hands. In the first proposed approach \citep{wiggers1}, we applied a pre-trained CNN as feature extractor considering feature maps of different sizes and explored the idea of obtaining a compact representation to provide a fast image retrieval. Thus, by transfer learning, we succeeded in obtaining a robust representation, while reducing the importance of the training phase which was represented in this case by a fine-tuning considering data augmentation on a small set of image queries. Going deeper into the idea of avoiding training based on queries, in the second approach \citep{wiggers2}, two deep models (CNN based) were organized in a pre-trained Siamese Convolutional Neural Network (SCNN) for feature extraction and similarity estimation. In both approaches, a strategy for reducing the dimensionality of the feature maps generated by the CNNs was proposed with the aim of reducing the computational cost of the retrieval and storage processes. 

Here, we describe the main steps of each approach and compare them using a more robust experimental protocol which is composed of two datasets, Tobacco800 \citep{LogoDetection-ICDAR07}, and DocExplore \citep{base}. In addition, the proposed approaches are compared to the state-of-the-art by addressing both image retrieval and pattern spotting tasks. The former task consists of finding all the document images that contain a given query, while in the latter, additional information is expected, which is related to the exact location of the query within the retrieved images. 
The contributions of this paper on the topics of image retrieval and pattern spotting on document images are threefold: (i) a method based on deep learning that explores the concepts of transfer learning and CNN fine-tuning to create representations (feature vectors) of different sizes, showing their impact in terms of precision and computation time; (ii) a method that explores deep SCNN models for image representation and similarity estimation which is query-independent, i.e. queries are not used in the off-line phase (training); (iii) we show that our flexible approaches based on deep models reach competitive results when compared to the state-of-the-art using a robust experimental protocol.

The remainder of this paper is organized as follows: Section 2 presents the proposed methods: the representation learning and each step of the image retrieval and pattern spotting tasks. Section 3 presents the Tobacco800 and DocExplore datasets used for the benchmark, followed by the experimental protocol and results.  Finally, Section 4 presents our conclusions and future work.

\section{Proposed Methods}
Image retrieval and pattern spotting are the main tasks of an image search engine. Several techniques can be used to perform such a search but they usually follow the same conventional process: (i) the document images are indexed and stored in an off-line phase; (ii) during an on-line phase, a similarity measure is used to compare query images with the stored documents, returning a ranked list of candidate documents. 

\begin{figure}[htpb!]
\centering
\includegraphics[width=0.48\textwidth]{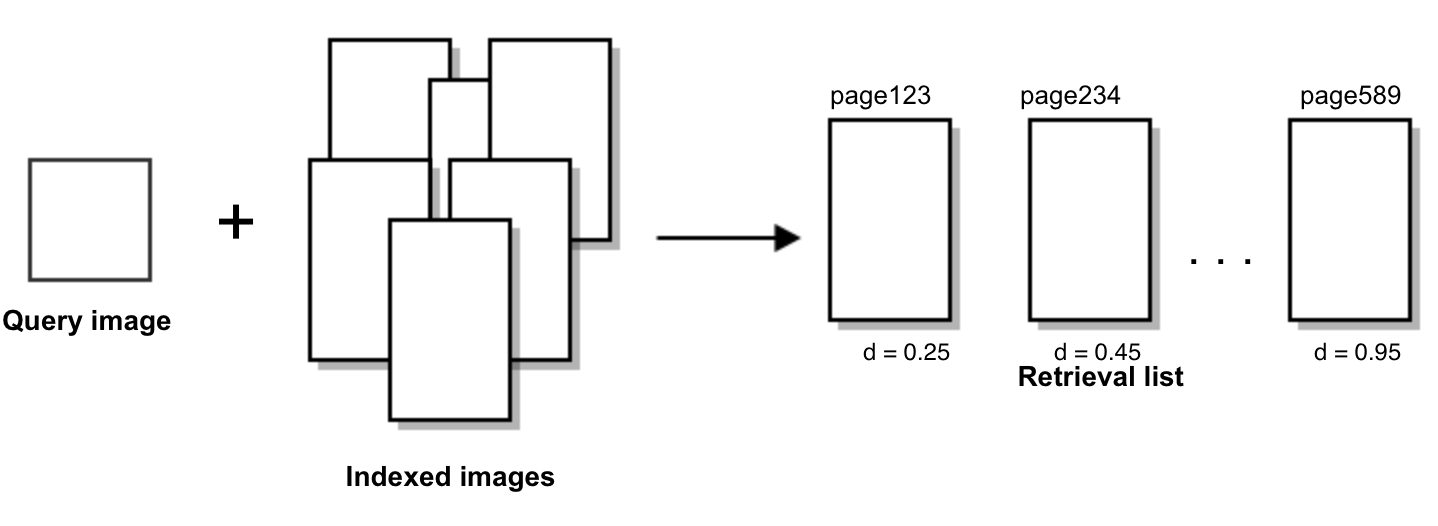}
\caption{Image Retrieval task returns a list of non-repeated images sorted by distance.}
\label{fig:ir}
\end{figure}

\begin{figure}[h]
\centering
\includegraphics[width=0.48\textwidth]{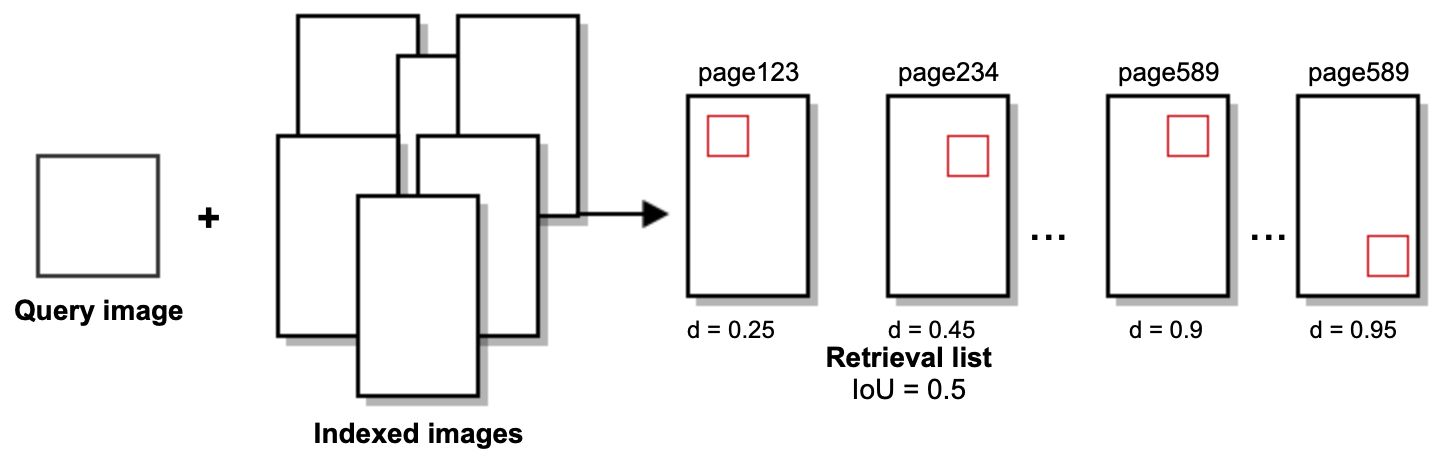}
\caption{Pattern Spotting task returns a list of images sorted by distance and the query location.}
\label{fig:ps}
\end{figure}

Whereas in the image retrieval task, an exhaustive search based on a distance measure is used to provide a rank with the Top-$k$ candidates that may contain the query (as shown in Fig.~\ref{fig:ir}), in the pattern spotting task the pattern (query) location(s) must be additionally provided on each document image as shown in Fig.~\ref{fig:ps} (red square). 

Despite the task, image retrieval or pattern spotting, three common steps are usually needed: a) pre-processing, b) representation (feature extraction), and c) retrieval (matching). We describe in this section the two approaches that share the same pre-processing step but differ in the ways the representation is obtained (considering or not contextual information) and how the similarity between the query and the image candidates is estimated (learned or non-learned metric).  

\subsection{Approach Based on CNN}
\begin{figure}[htbp]
\centering
\includegraphics[width=1.0\linewidth]{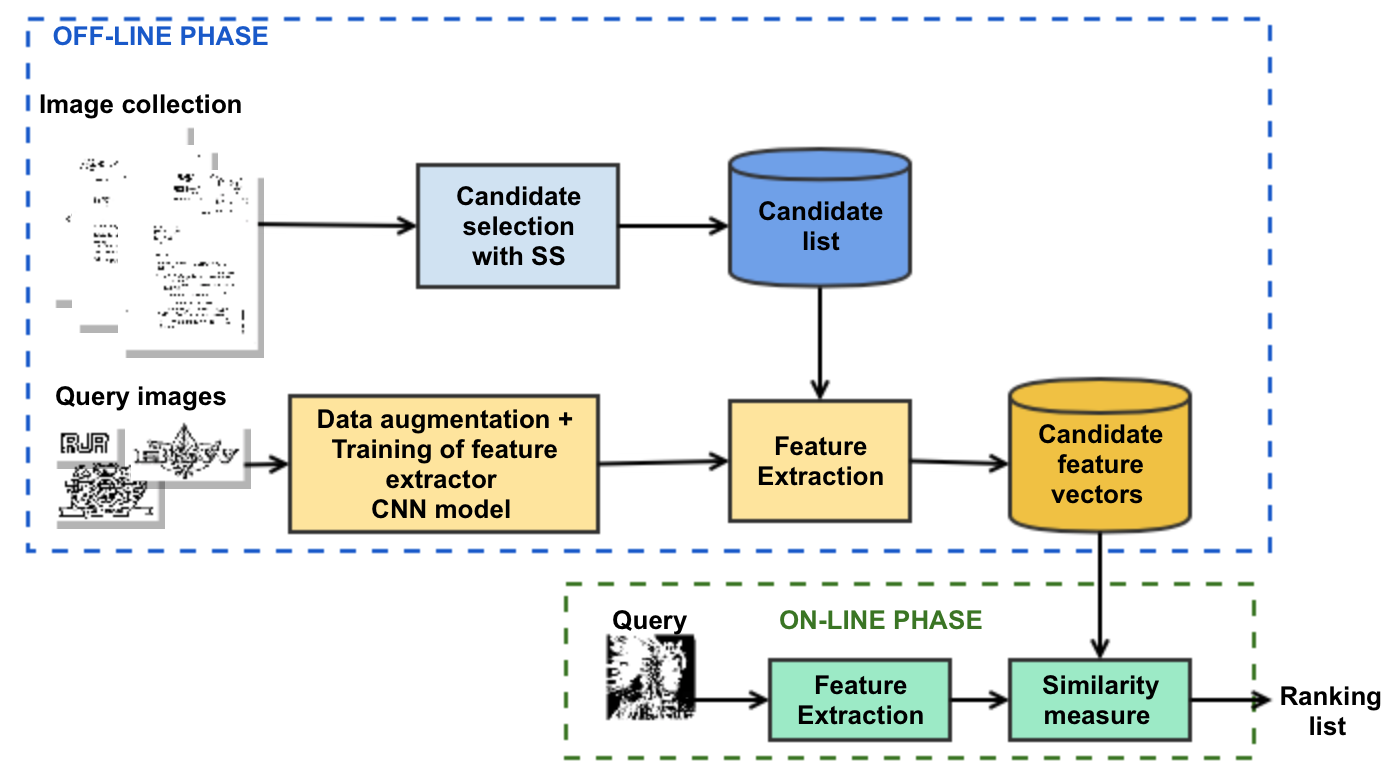}
\\
\caption{Overview of the CNN approach \citep{wiggers1}}
\label{fig:visao}
\end{figure}

 Fig.~\ref{fig:visao} shows an overview of the first proposed approach. In the off-line phase, given the document images, the Selective Search (SS) proposed by \cite{Uijlings2013} is used as a pre-processing step that combines the use of an exhaustive search with segmentation methods. We did not modify the parameters of the SS algorithm, which means that the limits for object location were kept as recommended in \citep{Uijlings2013}. More details about this implementation can be found in \cite{wiggers1}.

For extracting features from the queries and the candidate regions, we use the concept of transfer learning, i.e. a CNN model pre-trained on the ImageNet dataset is used as a feature extractor. The CNN model is based on the AlexNet architecture \citep{alex}. Note that there are many network topologies that could have been used also, for example, \cite{goingdepper} proposed GoogLenet, but their model is theoretically 50\% more complex than AlexNet; 
 \cite{simon} proposed an architecture with very small convolution filters, which is slower to train and needs more memory. Therefore, the AlexNet has been considered as a good starting point for the proposed approach. It is easy to implement and has shown to be effective in different deep learning scenarios, showing a good compromise between the number of layers and final accuracy. In addition, the non-linear part trains much faster than standard functions (sigmoid and tanh) and it achieved promising results on the ImageNet challenges.

 The AlexNet is made up of five convolutional layers followed by three fully connected layers \citep{alex}. An additional layer was introduced in the network \citep{7301269} to reduce the dimensionality of the feature maps provided at the output of the last convolutional layer. The feature map generated by the trained model is used to represent both the query and the candidate regions. We have collected the query images and performed data augmentation (40$\times$ the number of queries) to fine-tune the AlexNet. The feature map of each candidate is stored to be further used in the on-line phase. For the search process carried out during the on-line phase, given a query image, an exhaustive search is carried out considering the whole list of generated candidates. We have chosen the cosine distance because it is cheap to compute and according to \citet{en}, there is no statistical difference in performance with Euclidean distance. 
 
One main disadvantage of this approach is that the queries are used for training (fine-tuning) the CNN. Therefore, previous knowledge of the queries is needed a priori making this approach problem (or dataset) dependent. 
 
\subsection{Approach Based on Siamese Convolutional Neural Network (SCNN)}
\begin{figure}[htpb!]
\centering \includegraphics[width=1.0\linewidth]{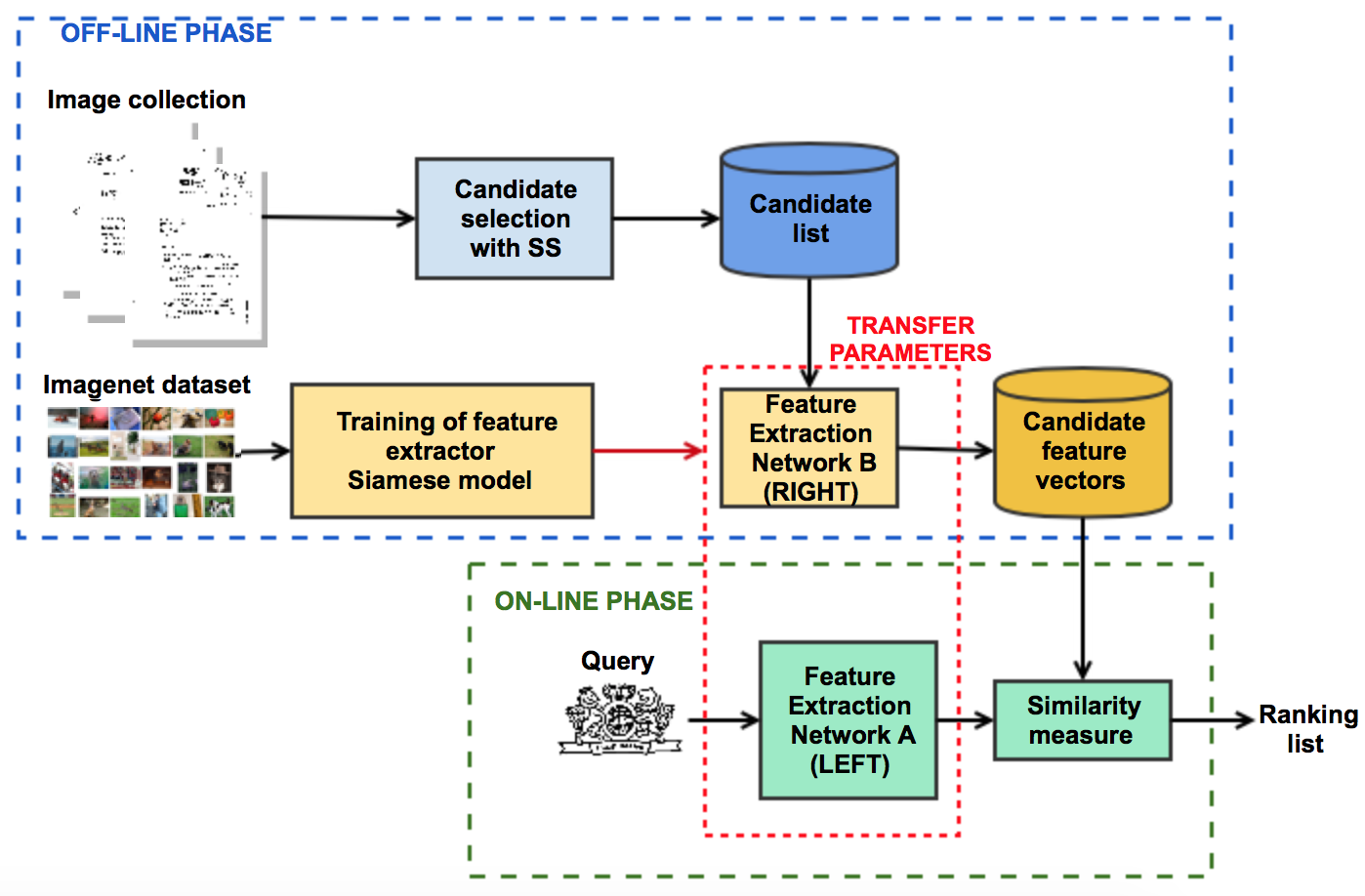}
\\
\caption{Overview of the SCNN approach \citep{wiggers2}}
\label{fig:overview}
\end{figure}

The feature vectors can be obtained by training Siamese neural networks using pairs of images \citep{10.1007/978-3-319-10590-1_38}. The Siamese model has been successfully used for face recognition \citep{facenet,8302003}, and signatures or symbol identification \citep{Koch2015SiameseNN}. In an SCNN, after the convolutional layer, a distance measure between vectors is computed according to two different activation samples during training. It can be seen as using two identical, parallel neural networks sharing the same set of weights \citep{BERLEMONT201847}. Fig.~\ref{fig:overview} presents a general overview of the proposed SCNN approach, that shares the same SS algorithm with the CNN approach. 

Inspired by \cite{7899663}, and \cite{chung}, and based on our first approach, the proposed SCNN architecture is composed of two identical AlexNet trained on the ImageNet dataset. The output layer was removed and the fully connected layer with $n$ dimensions was trained for learning a feature representation. Based on our previous work \citep{wiggers1}, we have considered an additional layer in the AlexNet architecture to evaluate feature maps of different sizes. 

Different from the conventional SCNN which uses contrastive loss \citep{chung,7299135}, we implemented Euclidean distance with the proposed layers available in the Caffe tool \citep{jiacaffe}, as follows: {\it Eltwise} to compute the distance between the feature map entries; {\it Power} to compute the square root; and {\it Reduction} to sum the obtained distances reducing it to a scalar. After that, we implement a fully connected layer with one output and a loss layer. The proposed SCNN architecture has the distance layers isolated from the loss layer. It may facilitate the use of the distance value in the on-line phase of the method. We set the learning rate to $10^{-3}$ with an exponential function. The activation function used was Stochastic Gradient Descent, as recommended by \cite{NIPS2011_0485}. 

During the training, the input is a pair of images $x\in X$ and $y\in Y$, where $X$ and $Y$ are ImageNet positive and negative pairs in the training set, respectively. The inputs $x$ and $y$ are fed into the two AlexNet networks $A$ and $B$, both with an additional layer with $n$ dimensions. The images of ImageNet used to train the SCNN were 100,000 similar pairs and 150,000 non-similar pairs. We generated 1.5$\times$ more non-similar images for training and test as recommended by \cite{7899663}. We split the images into training (70\%) and test (30\%) subsets. The output distance of $A$ and $B$ is fed to a \textit{sigmoid cross entropy loss} layer, that aims to minimize the difference of the probability distribution between the predicted labels and ground truth \citep{7926727}. 

During the testing step, the previously trained SCNN network is applied to each pair (query, region candidate) and returns the distance between the query and the region candidate learned by the SCNN through the distance layer implemented in the model \cite{wiggers2}. The retrieved document images are then ranked according to this distance.

The main advantage over the previous approach is that we do not need to know the queries in advance, i.e. during the off-line phase (training), thus being independent from the queries. Therefore, the SCNN based approach may be viewed as a problem (or dataset) independent approach for retrieving document images and spotting patterns, as we will demonstrate in the experiments (Section 3) by using the same trained SCNN to retrieve images from two very different datasets.   

\section{Experimental Results and Discussion}

\subsection{Datasets}
Two datasets were considered for evaluating the proposed approaches and they include a great variability of page layouts, being grayscale and color documents. \textit{Tobacco800 dataset} \citep{LogoDetection-ICDAR07} is a public subset of the Complex Document Image Processing (CDIP) collection, created by the Illinois Institute of Technology \citep{lewis}. It contains 412 document images with logos and 878 without logos. In our experiments, we have considered the 21 categories presenting two or more occurrences, making 418 queries for the search process. This dataset was used for evaluating the two approaches, i.e. CNN and SCNN-based. The SS algorithm was applied at the pre-processing stage in all the documents of the dataset and it generated 1.8M candidate images to the CNN approach without modifying parameters \citep{wiggers1}. In a second experiment \citep{wiggers2}, we modified the parameters for object location, changing the limits for object location search. We implemented an adaptive threshold scheme to change the block size and offset (constant subtracted from the neighborhood mean) values. Thus, the parameters were defined as follows: {\it block}$=$241 and {\it offset}$=$0.12, {\it k}$=$50 and 100, {\it color+texture+filler+size}. With this second configuration, we observed an increase (13$\times$ better as reported in \citep{wiggers2}) in the quality of the candidates that overlap the query in more than 90\%, when compared to our previous approach \citep{wiggers1}. This finally resulted in 1.2M candidate images that are used in the image retrieval and pattern spotting methods, for both the CNN and SCNN approaches. 

The second dataset is \textit{DocExplore dataset}\footnote{Available at \url{http://spotting.univ-rouen.fr}} from the Project DocExplore \citep{base}. All manuscripts date from the 10th through the 16th century. The 1500 images available in the dataset are organized into 35 categories and 1447 objects. The number of occurrences of each object in the collection can vary from two to more than 100. The objects differ in color, shape, size, distortion and possible degradation of the manuscript. This dataset was used only with the SCNN approach because it enables us to evaluate the concept of transfer learning in exactly the same conditions as \citep{base} for fair comparison, i.e. without training our system from a set of queries. The second configuration of SS was also applied to the document images using adaptive threshold with {\it block}$=$209 and  {\it offset}$=$0.14, {\it k}$=$50, 100 and 150; feature space RGB and normalize RGB; {\it color+fill+size+texture} and we reduced the number of candidates from 45M to 36M.

For evaluating the performance of the image retrieval task, Average Precision (AP) and Recall for each query are adopted as performance measures. For evaluating the performance of the pattern spotting task, the overlap between the query and the candidate (region of interest) retrieved must be computed. For this purpose, the Intersection over Union (IoU) is computed: let ($x_q$,$y_q$) denote the position of the ground truth in the document image corresponding to the query, $q$ its corresponding area, ($x_r$,$y_r$) the position of the candidate retrieved, $o$ its corresponding area, the candidate region is considered as relevant if its IoU as measured by Equation \ref{eq:iou} is above a given threshold \citep{sebastian}. 

\begin{equation}
IoU(x,y)= \frac{q\cap o}{q\cup o}
\label{eq:iou}
\end{equation}

In the following, we consider the analysis with 0.1$\leq$IoU$\leq$0.7 in order to determine that a positive candidate is retrieved, and to the end, the precision and the recall are calculated. Finally, the mAP is calculated to evaluate the results considering all the queries.

\subsection{Results on Tobacco800 Dataset}
Table \ref{table:twowig} shows the experimental results comparing the CNN \citep{wiggers1} and the SCNN \citep{wiggers2} approaches with feature maps of different sizes (4096, 512, 256, 128). As one may see, for Top-10 the SCNN-4096 and SCNN-512 is 28.23\% and 20.75\% better than the CNN approach, respectively. However, considering the maps with 256 and 128 features, the results are 9.20\% and 5.07\% worst, respectively. We observe that the SCNN is more discriminant with high dimensional features while the CNN returns better performance with low dimensional features, though the difference between the approaches with 128 features is only 0.03 of mAP in Top-10. Thus, the best choice depends on the required computational cost and the peculiarities of the dataset to achieve promising results.

\begin{table}[htpb!]
\small
        \centering
\caption{Comparison of the two proposed approaches in terms of mAP for Image Retrieval. Top-\textit{k} ranking for 4096, 512, 256 and 128 features on Tobacco800 dataset.}
        \label{table:twowig}
        \begin{tabular}{l|  c | c  | c | c | c  }
        \hline
       \multirow{2}{*}{\textbf{Method}} & {\textbf{Feature Map}} &  \multicolumn{4}{c}{\bf Top-\textit{k}}  \\ \cline{3-6}
                                & {\textbf{Dimension}} & \textbf{10} & \textbf{25} & \textbf{50}  & \textbf{100}\\ \hline
 
       \multirow{4}{*}{CNN} & 4096  &0.68 &0.57 &0.45  &0.32 \\ 
               &  512 & 0.53 & 0.40 & 0.31 & 0.23 \\
         &256 &\textbf{0.72}  &\textbf{0.61} &0.50 & \textbf{0.35} \\ 
         &128 & 0.69& 0.61& \textbf{0.51}& 0.34\\
         \hline
          \multirow{4}{*}{SCNN }& 4096  & \textbf{0.87}  &\textbf{0.73} &\textbf{0.54} &\textbf{0.30} \\
        & 512  &0.64 &  0.48&  0.35& 0.22\\
        & 256 &0.65  & 0.50& 0.36&0.22 \\
        & 128  & 0.65 & 0.50 & 0.37 & 0.22 \\
    \hline
        \end{tabular}
\end{table}

In addition, to evaluate the quality of candidates using the improved version of SS, we did experiments with CNN-4096 and improved SS of Tobacco800 dataset: the Top-10 ranking returned 0.75 of mAP, thus we increase 0.07 mAP (before was 0.68) but still it is 14\% below the improved SS with SCNN-4096. Then the use of new SS improved the results, but we get better results using new SS with SCNN.

Table \ref{table:tableiou} shows the experimental results of the pattern spotting task. Here, the candidate is considered as relevant if it overlaps enough with the image query. We can observe that the highest mAP (0.75) was achieved using the SCNN-4096, but the results related to the reduced maps are quite competitive, mainly for the CNN-256. Despite the reduction in mAP, they have shown a significant reduction in computational time. Table \ref{table:tableiou} also shows a small gap in terms of localization performance between IoU$=$0.1 and IoU$=$0.7. This means that our approach succeeded not only in retrieving the relevant image candidates for each query but also in finding the query position precisely.

\begin{table}[htpb!]
\vspace{-10pt}
\small
\centering
\caption{Pattern spotting results (mAP) considering different values of IoU and Top-10 on Tobacco800 dataset.}
\label{table:tableiou}
\begin{tabular}{l|c|c|c|c|c}
\hline
 \multirow{2}{*}{\textbf{Method}}  &  \bf Feature Map &  \multicolumn{4}{c}{\textbf{IoU}}  \\ \cline{3-6}
  & \bf Dimension & \textbf{0.1} & \textbf{0.3} & \textbf{0.5}  & \textbf{0.7} \\ \hline
  
\multirow{4}{*}{CNN} & 4096 & 0.65 & 0.65  & 0.64 & 0.61\\
&512 &0.50  & 0.50& 0.50 &0.49\\
&256 & \textbf{0.69} & \textbf{0.69} & \textbf{0.60} & \textbf{0.56}\\
&128 &  0.67& 0.67& 0.60&0.59\\ 
  \hline
 \multirow{4}{*}{SCNN} & 4096 & \textbf{0.75} & \textbf{0.75}  &\textbf{0.74}  & \textbf{0.74}   \\
& 512 & 0.61  & 0.61 & 0.61 & 0.60\\
& 256& 0.63 & 0.63 &0.63 & 0.62 \\
& 128 & 0.63 & 0.63 & 0.62 & 0.62 \\
\hline
\end{tabular}
\end{table}

\begin{figure}[h!]
\centering
\includegraphics[width=0.45\textwidth]{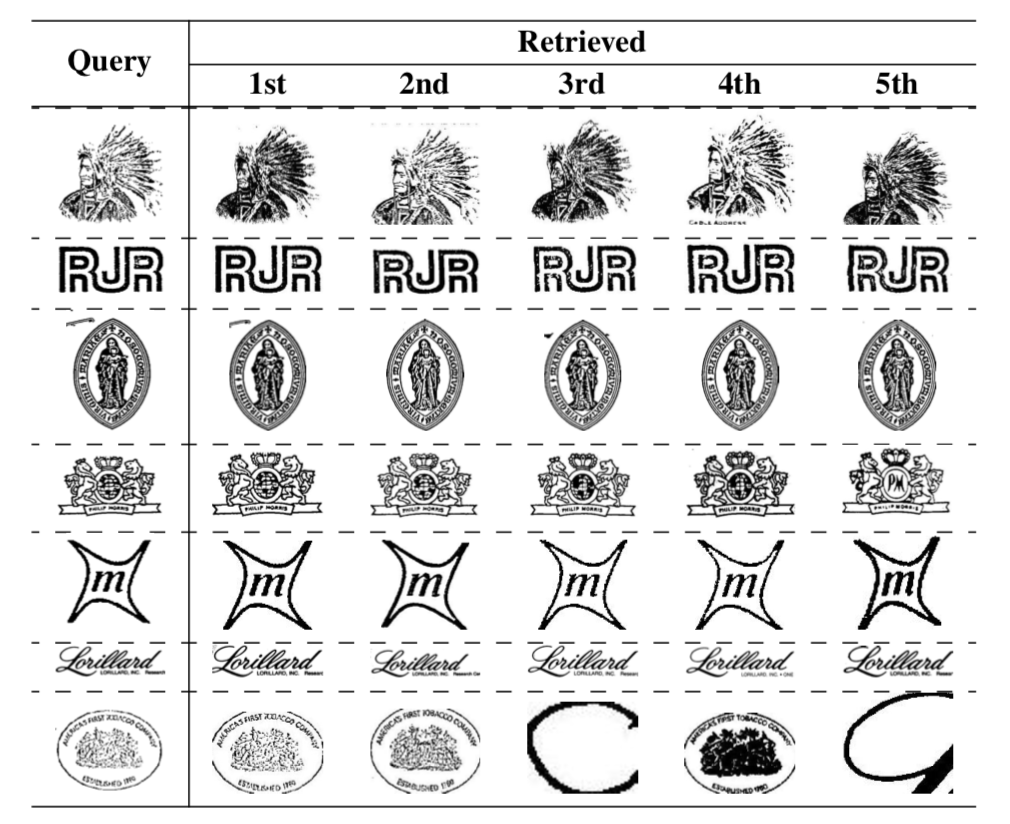}
\caption{Qualitative retrieval results for Tobacco800 dataset. Query and the first five retrieved logos using 4096 features}
\label{fig:qualitative}
\end{figure}

Fig.~\ref{fig:qualitative} shows some qualitative results of the logos retrieved using the SCNN-4096 because of better performance in all the experiments. These results are very promising since many correct logos were retrieved. We can observe the good performance especially in the sixth row, where the query is very similar to a signature, but we did not have false positives. In the last row we can see some false candidates, because of the similarity of the edges along with the presence of too few positive samples in the dataset.

The results in image retrieval were compared with the current state-of-the-art, and the results are shown in Table \ref{table:tablerus}. The proposed CNN approach is 51\% and 60\%  better than the mAP reported by \cite{jain2} considering the 4096 and 256 features, respectively. On the other hand, the mAP achieved by the CNN-256 is 15\% below than the approach proposed by \cite{rusinolTobacco}. This means that the candidates retrieved by the proposed approach are not well ranked in the first positions or that there are too many false positives. The SCNN approach is 93.80\% and 45.33\% better than the results achieved by \cite{jain2}, with 4096 and 256 features, respectively. The SCNN-4096 is 2.08\% better the results achieved by \cite{rusinolTobacco}. However, the results are almost 30\% below for model with 256 features. 

\begin{table}[htpb!]
\vspace{-10pt}
\small
\begin{center}
\caption{Comparison with the state-of-the-art for retrieval task (mAP). The CNN and the SCNN-based methods, considering Top-10 and Top-25 on Tobacco800 dataset.}
\label{table:tablerus}
\begin{tabular}{l| c |c  }
\hline
\multirow{2}{*}{\bf Method}& \multicolumn{2}{c}{\bf Top-\textit{k}}\\ \cline{2-3}
       & \textbf{10} & \textbf{25}\\  
                              \hline
\cite{jain2} & 0.45 & NA  \\
  \cite{rusinolTobacco} & NA &  0.72\\ \hline
 
 CNN-4096 & 0.68 &0.57 \\
 SCNN-4096 & \textbf{0.87} & \textbf{0.73} \\ \hline
 CNN-256 &0.72 & 0.61 \\
 SCNN-256 &0.65 & 0.50 \\  
\hline
\multicolumn{3}{l}{\scriptsize NA: Not Available} 
\end{tabular}
\end{center}
\vspace{-15pt}
\end{table}


Table \ref{table:state} shows a comparison with the current state-of-the-art for pattern spotting on the Tobacco800 dataset. We selected our best image retrieval results which have been obtained with the SCNN-4096 and the CNN-256 to evaluate the PS task. For such a comparison we have used the same experimental parameters of \cite{phuong,6628626}, who consider Top-5, IoU$\geq$0.6 and classes with at least three samples per category. The mAP achieved by the CNN-256 did not outperform the state of the art. However, the mAP achieved by the proposed SCNN-4096 is 1.31\% better than mAP achieved by \cite{phuong} but it did not outperform the mAP presented in \cite{6628626}. On the other hand, the recall is 4.44\% and 4.86\% better than \cite{phuong} and \cite{6628626} respectively. It is important to highlight that both \cite{6628626} and \cite{phuong} require the previous knowledge of the logo gallery, which is not necessary for the proposed method. 

\begin{table}[htpb!]
\vspace{-10pt}
\small
\begin{center}
\caption{Comparison with the state-of-the-art for Pattern Spotting in terms of mAP. Top-5, IoU$\geq$0.6, and classes with at least three samples of Tobacco800 dataset.}
\label{table:state}
\begin{tabular}{l|c|c}
\hline
\multirow{1}{*}{\bf Method} & \multirow{1}{*}{\bf mAP} & \bf Recall (\%) \\ 
\hline
 \cite{6628626}	& \textbf{0.970} & 88.42 \\
\cite{phuong} & 0.910 & 88.78 \\	
SCNN-4096 &  0.922 & \textbf{92.72} \\
CNN-256 & 0.600 & 63.00 \\
\hline			     
\end{tabular}
\end{center}
\vspace{-10pt}
\end{table}

We can therefore conclude that the results related to the reduced maps are quite competitive. Besides, the computation time is another very important performance measure for retrieval systems. There is not a significant difference in the computation time between CNN and SCNN, since the on-line phase depends on feature maps already extracted to calculate the distances and build the ranking list of both models. During the on-line phase, in both models, the query search time is improved from $\cong7$s (4096-feature map) to $\cong3$s (256-feature map). Therefore, the computation time decreases when using the model with 256 features, while the approaches return promising results.

\subsection{Results on DocExplore Dataset}
Table \ref{table:docresults} shows the experimental results of the proposed image retrieval and pattern spotting methods, using the SCNN-4096. We selected this feature map because of its performance in Tobacco800 dataset. We evaluated our results thanks to the evaluation kit provided on-line\footnote{\url{http://spotting.univ-rouen.fr/evaluation-kit/}}, where the authors included a rule to ignore junk objects and IoU$\geq$0.5 for PS task. We can observe that our best result for IR is with Top-1000, but for Top-500 the difference is only 3.5\%. For PS our results are very similar to any Top-k ranking, but in our best result we achieved 0.076 of mAP.  

As \cite{7899938}, we observed that many candidates often contain only a part of the query or overlap with other ranked candidates, hence reducing the performance of the system. Thus, we propose a post-processing stage to use a "union" of these retained candidates to discover rectangular regions as a way to improve the performance of the localization task. Thus, we selected the first 2000 candidates to apply the union step. After the union, the first 1000 were considered to feed the evaluation system, similar to \cite{base} and \cite{7899938}. We can observe that our results for PS are approximately 2.3 times better using post-processing.

\begin{table}[htpb!]
\small
        \centering
\caption{Image retrieval (IR) and Pattern spotting (PS) results with Top-\textit{k} ranking for SCNN-4096 on DocExplore dataset. 
}
        \label{table:docresults}
        \begin{tabular}{l| c |c  | c | c | c  }
        \hline
       \multirow{2}{*}{\textbf{Task}}  &  \multicolumn{5}{c}{\bf Top-\textit{k}}  \\ \cline{2-6}
                               & \textbf{100} & \textbf{300} & \textbf{500}  &\textbf{700} & \textbf{1000}\\ \hline
       IR & 0.296 &0.355 &0.373 & 0.381&  \bf 0.386 \\ 
       PS & 0.073 & 0.075& 0.076& 0.076& 0.076 \\
       PS (PP)& 0.174 & 0.174 & 0.174 & 0.174 & \bf 0.174\\
      \hline
      \multicolumn{6}{l}{\scriptsize PP: Post-processing}
        \end{tabular}
\vspace{-10pt}
\end{table}

Table \ref{table:results_docIR1} shows the comparison with the current state-of-the-art and our experimental results. 
It can be observed that we did not outperform \cite{base}, being almost 37\% worse for IR. They used BING to propose regions and the feature extraction is based on VLAD and K-means. In addition, the authors included a post-processing stage using template matching. At each retrieval, the Top-2000 regions similar to the query are kept as inputs of the template matching stage. Finally, only the first 1000 matched regions are used for calculating the mAP. It can be observed that the mAP achieved by the proposed method in the localization task outperforms in 57\% the results presented in \citep{base} when considering our post-processing method.

\begin{table}[htpb!]
\vspace{-10pt}
\small
        \centering
\caption{Image retrieval (IR) and pattern spotting (PS) results on DocExplore dataset. Values refer to mAP on Top-1000 candidates for SCNN-4096}
        \label{table:results_docIR1}
\begin{tabular}{l|c|c} \hline
\bf Method      &\bf IR & \bf PS \\  \hline
                    
\cite{base}  & \bf 0.613  &  0.111            \\
\cite{chile}  ES  & 0.286 &  0.139  \\
\cite{chile}  PP  & 0.386 &   0.173           \\ 
SCNN-4096 PP & 0.386 & \bf 0.174 \\
\hline
\multicolumn{3}{l}{\scriptsize ES: exhaustive search,  PP: Post-processing}
\end{tabular}
\vspace{-10pt}
\end{table}
    
The mAP for IR achieved by the proposed SCNN is similar to the mAP achieved by \cite{chile} with post-processing (PP). \cite{chile} used RetinaNet as a feature extractor and added a post-processing step where they discard the localization if the bounding box is not entirely contained in the original page. In addition, they trained a classifier with a set of pages of DocExplore and use it to predict non-text regions in all pages to generate region proposals. It is important to notice that we did not perform any training using images of DocExplore for both the region proposal and the feature extractor. However, whereas the authors used an exhaustive search, our results outperformed in approximately 35\% in IR and 25\% for PS. Therefore, our approach is better at retrieving patterns and it does not need any additional classifier to improve the performance.

\begin{figure}[h!]
\centering
\includegraphics[width=0.48\textwidth]{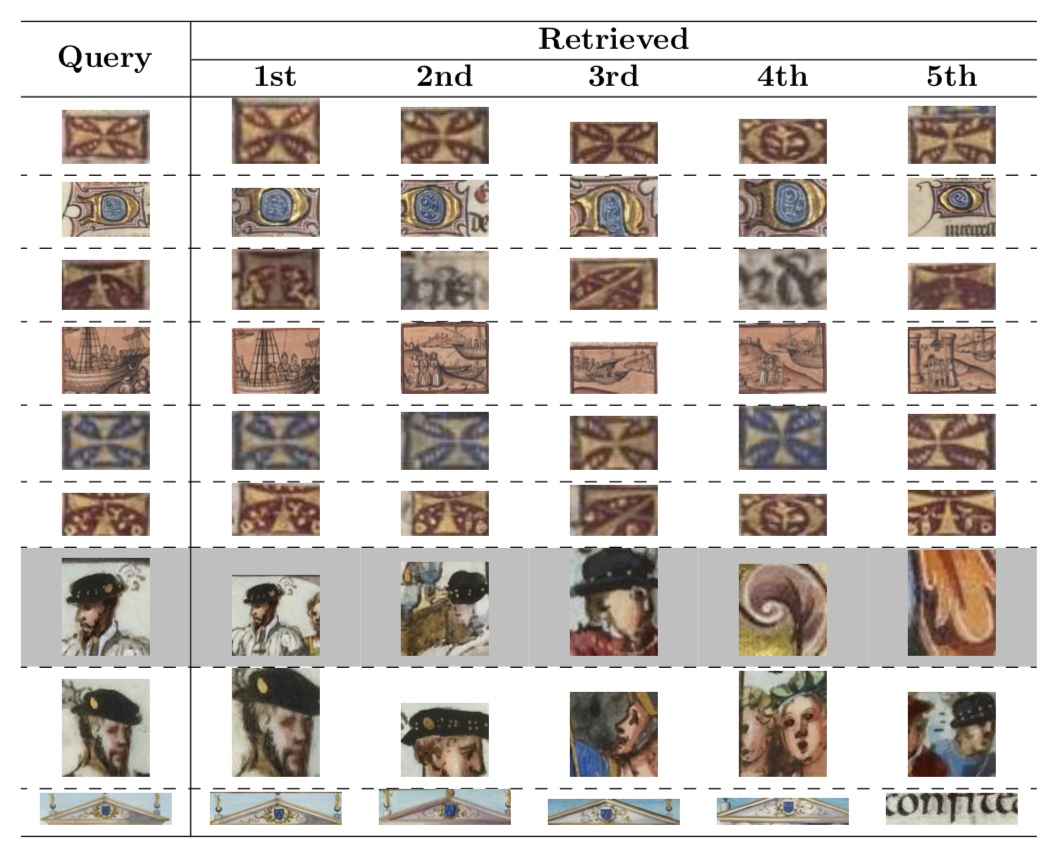}
\caption{Qualitative retrieval results DocExplore dataset. Query and the first five retrieved logos using SCNN-4096}
\label{fig:qualitativedocexplore}
\vspace{-5pt}
\end{figure}

Fig.~\ref{fig:qualitativedocexplore} shows some qualitative results of the images retrieved using SCNN-4096. These results are very promising, however, when the search involves an image with few positive samples, the method returns some false candidates, as seen in the seventh row (in gray), although the level of detail of the candidate retrieved in the first position is very high. The same occurs in the eighth row. 

\section{Conclusion}
We presented two approaches for retrieving images and spotting patterns in document image collections using deep learning. In the first approach a pre-trained AlexNet model was used to cope with the lack of training data. The CNN was fine-tuned to achieve a compact yet discriminant representation of queries and image candidates. A constraint of this approach is however the need to fine-tune the CNN, making it a dataset dependent solution. We solved it in the second approach, in which a SCNN trained on a subset of the ImageNet dataset avoids the fine-tuning process, making it independent from the queries and the dataset used. 

A robust experimental protocol using two public datasets (Tobacoo-800 and DocExplore) has shown that the proposed methods compare favorably against state-of-the-art document image retrieval and pattern spotting methods. Though in both approaches the learned representation scheme considers feature maps of different sizes which were evaluated in terms of retrieval performance, it was possible to observe with the similarity-based features provided by the SCNN with a 4096 feature map an increase in the mAP since the features generalize better and allow to improve the matching performance. 

Further work must be done to evaluate other deep architectures and the possible fusion of the proposed learned representation with the handcrafted features described in \citep{en} that are very discriminating and competitive. 

\section*{Acknowledgments}
\noindent CAPES (Coord. for the Improvement of Higher Education Personnel) and CNPq (National Council for Scientific and Technological Development) grant 306684/2018-2. 

\bibliographystyle{model2-names}

\begin{thebibliography}{37}
\expandafter\ifx\csname natexlab\endcsname\relax\def\natexlab#1{#1}\fi
\providecommand{\url}[1]{\texttt{#1}}
\providecommand{\href}[2]{#2}
\providecommand{\path}[1]{#1}
\providecommand{\DOIprefix}{doi:}
\providecommand{\ArXivprefix}{arXiv:}
\providecommand{\URLprefix}{URL: }
\providecommand{\Pubmedprefix}{pmid:}
\providecommand{\doi}[1]{\href{http://dx.doi.org/#1}{\path{#1}}}
\providecommand{\Pubmed}[1]{\href{pmid:#1}{\path{#1}}}
\providecommand{\bibinfo}[2]{#2}
\ifx\xfnm\relax \def\xfnm[#1]{\unskip,\space#1}\fi
\bibitem[{Babenko and Lempitsky(2015)}]{Babenko_2015_ICCV}
\bibinfo{author}{Babenko, A.}, \bibinfo{author}{Lempitsky, V.},
  \bibinfo{year}{2015}.
\newblock \bibinfo{title}{Aggregating local deep features for image retrieval},
  in: \bibinfo{booktitle}{IEEE Intl Conf Comp Vision}.
\bibitem[{Babenko et~al.(2014)Babenko, Slesarev, Chigorin and
  Lempitsky}]{10.1007/978-3-319-10590-1_38}
\bibinfo{author}{Babenko, A.}, \bibinfo{author}{Slesarev, A.},
  \bibinfo{author}{Chigorin, A.}, \bibinfo{author}{Lempitsky, V.},
  \bibinfo{year}{2014}.
\newblock \bibinfo{title}{Neural codes for image retrieval}, in:
  \bibinfo{booktitle}{Europ Conf Comp Vision}, pp. \bibinfo{pages}{584--599}.
\bibitem[{Berlemont et~al.(2018)Berlemont, Lefebvre, Duffner and
  Garcia}]{BERLEMONT201847}
\bibinfo{author}{Berlemont, S.}, \bibinfo{author}{Lefebvre, G.},
  \bibinfo{author}{Duffner, S.}, \bibinfo{author}{Garcia, C.},
  \bibinfo{year}{2018}.
\newblock \bibinfo{title}{Class-balanced siamese neural networks}.
\newblock \bibinfo{journal}{Neurocomputing} \bibinfo{volume}{273},
  \bibinfo{pages}{47 -- 56}.
\bibitem[{Chung and Weng(2017)}]{chung}
\bibinfo{author}{Chung, Y.A.}, \bibinfo{author}{Weng, W.H.},
  \bibinfo{year}{2017}.
\newblock \bibinfo{title}{Learning deep representations of medical images using
  siamese cnns with application to content-based image retrieval}, in:
  \bibinfo{booktitle}{NIPS 2017 Works Mach Learn for Health}.
\bibitem[{En et~al.(2016a)En, Nicolas, Petitjean, Jurie and Heutte}]{base}
\bibinfo{author}{En, S.}, \bibinfo{author}{Nicolas, S.},
  \bibinfo{author}{Petitjean, C.}, \bibinfo{author}{Jurie, F.},
  \bibinfo{author}{Heutte, L.}, \bibinfo{year}{2016}a.
\newblock \bibinfo{title}{New public dataset for spotting patterns in medieval
  document images}.
\newblock \bibinfo{journal}{Journal of Electronic Imaging}
  \bibinfo{volume}{26}.
\bibitem[{En et~al.(2016b)En, Petitjean, Nicolas and Heutte}]{en}
\bibinfo{author}{En, S.}, \bibinfo{author}{Petitjean, C.},
  \bibinfo{author}{Nicolas, S.}, \bibinfo{author}{Heutte, L.},
  \bibinfo{year}{2016}b.
\newblock \bibinfo{title}{A scalable pattern spotting system for historical
  documents}.
\newblock \bibinfo{journal}{Patt Recog} \bibinfo{volume}{54},
  \bibinfo{pages}{149--161}.
\bibitem[{En et~al.(2016c)En, Petitjean, Nicolas, Heutte and Jurie}]{7899938}
\bibinfo{author}{En, S.}, \bibinfo{author}{Petitjean, C.},
  \bibinfo{author}{Nicolas, S.}, \bibinfo{author}{Heutte, L.},
  \bibinfo{author}{Jurie, F.}, \bibinfo{year}{2016}c.
\newblock \bibinfo{title}{Pattern localization in historical document images
  via template matching}, in: \bibinfo{booktitle}{23rd Intl Conf Patt Recog},
  pp. \bibinfo{pages}{2054--2059}.
\bibitem[{Gordo et~al.(2016)Gordo, Almaz{\'a}n, Revaud and Larlus}]{Gordo2016}
\bibinfo{author}{Gordo, A.}, \bibinfo{author}{Almaz{\'a}n, J.},
  \bibinfo{author}{Revaud, J.}, \bibinfo{author}{Larlus, D.},
  \bibinfo{year}{2016}.
\newblock \bibinfo{title}{Deep image retrieval: Learning global representations
  for image search}, in: \bibinfo{booktitle}{14th Europ Conf Comp Vision}, pp.
  \bibinfo{pages}{241--257}.
\bibitem[{Jain and Doermann(2012)}]{jain2}
\bibinfo{author}{Jain, R.}, \bibinfo{author}{Doermann, D.},
  \bibinfo{year}{2012}.
\newblock \bibinfo{title}{Logo retrieval in document images}, in:
  \bibinfo{booktitle}{10th IAPR Intl Works Doc Anal Syst}, pp.
  \bibinfo{pages}{135--139}.
\bibitem[{Jia et~al.(2014)Jia, Shelhamer, Donahue, Karayev, Long, Girshick,
  Guadarrama and Darrell}]{jiacaffe}
\bibinfo{author}{Jia, Y.}, \bibinfo{author}{Shelhamer, E.},
  \bibinfo{author}{Donahue, J.}, \bibinfo{author}{Karayev, S.},
  \bibinfo{author}{Long, J.}, \bibinfo{author}{Girshick, R.},
  \bibinfo{author}{Guadarrama, S.}, \bibinfo{author}{Darrell, T.},
  \bibinfo{year}{2014}.
\newblock \bibinfo{title}{Caffe: Convolutional architecture for fast feature
  embedding}.
\newblock \bibinfo{journal}{arXiv preprint arXiv:1408.5093} .
\bibitem[{Koch et~al.(2015)Koch, Zemel and Salakhutdinov}]{Koch2015SiameseNN}
\bibinfo{author}{Koch, G.}, \bibinfo{author}{Zemel, R.},
  \bibinfo{author}{Salakhutdinov, R.}, \bibinfo{year}{2015}.
\newblock \bibinfo{title}{Siamese neural networks for one-shot image
  recognition}, in: \bibinfo{booktitle}{ICML 2015 Deep Learning Works}.
\bibitem[{Krizhevsky et~al.(2012)Krizhevsky, Sutskever and Hinton}]{alex}
\bibinfo{author}{Krizhevsky, A.}, \bibinfo{author}{Sutskever, I.},
  \bibinfo{author}{Hinton, G.E.}, \bibinfo{year}{2012}.
\newblock \bibinfo{title}{Imagenet classification with deep convolutional
  neural networks}, in: \bibinfo{booktitle}{Adv Neural Inf Proc Syst}.
\bibitem[{Le et~al.(2014)Le, Nayef, Visani, Ogier and Tran}]{phuong}
\bibinfo{author}{Le, V.P.}, \bibinfo{author}{Nayef, N.},
  \bibinfo{author}{Visani, M.}, \bibinfo{author}{Ogier, J.M.},
  \bibinfo{author}{Tran, C.D.}, \bibinfo{year}{2014}.
\newblock \bibinfo{title}{Document retrieval based on logo spotting using
  key-point matching}, in: \bibinfo{booktitle}{22nd Intl Conf Patt Recog}, pp.
  \bibinfo{pages}{3056--3061}.
\bibitem[{Le et~al.(2013)Le, Visani, Tran and Ogier}]{6628626}
\bibinfo{author}{Le, V.P.}, \bibinfo{author}{Visani, M.},
  \bibinfo{author}{Tran, C.D.}, \bibinfo{author}{Ogier, J.M.},
  \bibinfo{year}{2013}.
\newblock \bibinfo{title}{Improving logo spotting and matching for document
  categorization by a post-filter based on homography}, in:
  \bibinfo{booktitle}{12th Intl Conf Doc Anal Recog}, pp.
  \bibinfo{pages}{270--274}.
\bibitem[{Lecun et~al.(2010)Lecun, Kavakcuoglu and Farabet}]{lecun2010}
\bibinfo{author}{Lecun, Y.}, \bibinfo{author}{Kavakcuoglu, K.},
  \bibinfo{author}{Farabet, C.}, \bibinfo{year}{2010}.
\newblock \bibinfo{title}{Convolutional networks and applications in vision},
  in: \bibinfo{booktitle}{Proc of ISCAS}, \bibinfo{address}{Paris}. pp.
  \bibinfo{pages}{253--256}.
\bibitem[{Lewis et~al.(2006)Lewis, Agam, Argamon, Frieder, Grossman and
  Heard}]{lewis}
\bibinfo{author}{Lewis, D.}, \bibinfo{author}{Agam, G.},
  \bibinfo{author}{Argamon, S.}, \bibinfo{author}{Frieder, O.},
  \bibinfo{author}{Grossman, D.}, \bibinfo{author}{Heard, J.},
  \bibinfo{year}{2006}.
\newblock \bibinfo{title}{Building a test collection for complex document
  information processing}, in: \bibinfo{booktitle}{29th Annual Intl ACM SIGIR
  Conf}, pp. \bibinfo{pages}{665--666}.
\bibitem[{Lin et~al.(2015a)Lin, Yang, Hsiao and Chen}]{7301269}
\bibinfo{author}{Lin, K.}, \bibinfo{author}{Yang, H.F.},
  \bibinfo{author}{Hsiao, J.H.}, \bibinfo{author}{Chen, C.S.},
  \bibinfo{year}{2015}a.
\newblock \bibinfo{title}{Deep learning of binary hash codes for fast image
  retrieval}, in: \bibinfo{booktitle}{IEEE Conf Comp Vision Patt Recog}, pp.
  \bibinfo{pages}{27--35}.
\bibitem[{Lin et~al.(2015b)Lin, Cui, Belongie and Hays}]{7299135}
\bibinfo{author}{Lin, T.}, \bibinfo{author}{Cui, Y.},
  \bibinfo{author}{Belongie, S.}, \bibinfo{author}{Hays, J.},
  \bibinfo{year}{2015}b.
\newblock \bibinfo{title}{Learning deep representations for ground-to-aerial
  geolocalization}, in: \bibinfo{booktitle}{IEEE Conf on Comp Vision Patt
  Recog}, pp. \bibinfo{pages}{5007--5015}.
\bibitem[{Liu and Qi(2017)}]{7926727}
\bibinfo{author}{Liu, L.}, \bibinfo{author}{Qi, H.}, \bibinfo{year}{2017}.
\newblock \bibinfo{title}{Learning effective binary descriptors via cross
  entropy}, in: \bibinfo{booktitle}{IEEE Winter Conf Appl Comp Vision}, pp.
  \bibinfo{pages}{1251--1258}.
\bibitem[{Melekhov et~al.(2016)Melekhov, Kannala and Rahtu}]{7899663}
\bibinfo{author}{Melekhov, I.}, \bibinfo{author}{Kannala, J.},
  \bibinfo{author}{Rahtu, E.}, \bibinfo{year}{2016}.
\newblock \bibinfo{title}{Siamese network features for image matching}, in:
  \bibinfo{booktitle}{Intl Conf Patt Recogn}, pp. \bibinfo{pages}{378--383}.
\bibitem[{Nowozin(2014)}]{sebastian}
\bibinfo{author}{Nowozin, S.}, \bibinfo{year}{2014}.
\newblock \bibinfo{title}{Optimal decisions from probabilistic models: the
  intersection-over-union case}, in: \bibinfo{booktitle}{Comp Vision Patt Recog
  Conf}.
\bibitem[{Recht et~al.(2011)Recht, Re, Wright and Niu}]{NIPS2011_0485}
\bibinfo{author}{Recht, B.}, \bibinfo{author}{Re, C.}, \bibinfo{author}{Wright,
  S.}, \bibinfo{author}{Niu, F.}, \bibinfo{year}{2011}.
\newblock \bibinfo{title}{Hogwild: A lock-free approach to parallelizing
  stochastic gradient descent}, in: \bibinfo{booktitle}{Adv Neural Inf Proc
  Syst 24}, pp. \bibinfo{pages}{693--701}.
\bibitem[{Rusinol and Llad\'{o}s(2010)}]{rusinolTobacco}
\bibinfo{author}{Rusinol, M.}, \bibinfo{author}{Llad\'{o}s, J.},
  \bibinfo{year}{2010}.
\newblock \bibinfo{title}{Efficient logo retrieval through hashing shape
  context descriptors}, in: \bibinfo{booktitle}{9th IAPR Intl Works Doc Anal
  Syst}, pp. \bibinfo{pages}{215--222}.
\bibitem[{Schroff et~al.(2015)Schroff, Kalenichenko and Philbin}]{facenet}
\bibinfo{author}{Schroff, F.}, \bibinfo{author}{Kalenichenko, D.},
  \bibinfo{author}{Philbin, J.}, \bibinfo{year}{2015}.
\newblock \bibinfo{title}{Facenet: A unified embedding for face recognition and
  clustering}, in: \bibinfo{booktitle}{IEEE Conf Comp Vision Patt Recog}, pp.
  \bibinfo{pages}{815--823}.
\bibitem[{Simonyan and Zisserman.(2015)}]{simon}
\bibinfo{author}{Simonyan, K.}, \bibinfo{author}{Zisserman., A.},
  \bibinfo{year}{2015}.
\newblock \bibinfo{title}{Very deep convolutional networks for large-scale
  image recognition}, in: \bibinfo{booktitle}{5th Intl Conf Learn Repres}.
\bibitem[{Szegedy et~al.(2015)Szegedy, Liu, Jia, Sermanet, Reed, Anguelov,
  Erhan, Vanhoucke and Rabinovich}]{goingdepper}
\bibinfo{author}{Szegedy, C.}, \bibinfo{author}{Liu, W.}, \bibinfo{author}{Jia,
  Y.}, \bibinfo{author}{Sermanet, P.}, \bibinfo{author}{Reed, S.},
  \bibinfo{author}{Anguelov, D.}, \bibinfo{author}{Erhan, D.},
  \bibinfo{author}{Vanhoucke, V.}, \bibinfo{author}{Rabinovich, A.},
  \bibinfo{year}{2015}.
\newblock \bibinfo{title}{Going depper with convolutions}, in:
  \bibinfo{booktitle}{Comp Vision Patt Recog}.
\bibitem[{Thuy et~al.(2017)Thuy, Huu, Van and Quoc}]{DaoThiThuy201730}
\bibinfo{author}{Thuy, Q.D.T.}, \bibinfo{author}{Huu, Q.N.},
  \bibinfo{author}{Van, C.P.}, \bibinfo{author}{Quoc, T.N.},
  \bibinfo{year}{2017}.
\newblock \bibinfo{title}{An efficient semantic -- related image retrieval
  method}.
\newblock \bibinfo{journal}{Expert Sys App} \bibinfo{volume}{72},
  \bibinfo{pages}{30 -- 41}.
\bibitem[{\'Ubeda et~al.(2019)\'Ubeda, Saavedra, Nicolas, Petitjean and
  Heutte}]{chile}
\bibinfo{author}{\'Ubeda, I.}, \bibinfo{author}{Saavedra, J.M.},
  \bibinfo{author}{Nicolas, S.}, \bibinfo{author}{Petitjean, C.},
  \bibinfo{author}{Heutte, L.}, \bibinfo{year}{2019}.
\newblock \bibinfo{title}{Pattern spotting in historical documents using
  convolutional models}, in: \bibinfo{booktitle}{arXiv:1906.08580}.
\bibitem[{Uijlings et~al.(2013)Uijlings, van~de Sande, Gevers and
  Smeulders}]{Uijlings2013}
\bibinfo{author}{Uijlings, J.R.R.}, \bibinfo{author}{van~de Sande, K.E.A.},
  \bibinfo{author}{Gevers, T.}, \bibinfo{author}{Smeulders, A.W.M.},
  \bibinfo{year}{2013}.
\newblock \bibinfo{title}{Selective search for object recognition}.
\newblock \bibinfo{journal}{Intl J Comp Vision} \bibinfo{volume}{104},
  \bibinfo{pages}{154--171}.
\bibitem[{Wiggers et~al.(2018)Wiggers, Britto~Jr., Koerich, Heutte and
  Oliveira}]{wiggers1}
\bibinfo{author}{Wiggers, K.L.}, \bibinfo{author}{Britto~Jr., A.S.},
  \bibinfo{author}{Koerich, A.L.}, \bibinfo{author}{Heutte, L.},
  \bibinfo{author}{Oliveira, L.E.S.}, \bibinfo{year}{2018}.
\newblock \bibinfo{title}{Document image retrieval using deep features}, in:
  \bibinfo{booktitle}{Intl Joint Conf Neural Networks}, \bibinfo{address}{Rio
  de Janeiro}. pp. \bibinfo{pages}{3185--3192}.
\bibitem[{Wiggers et~al.(2019)Wiggers, Britto~Jr., Koerich, Heutte and
  Oliveira}]{wiggers2}
\bibinfo{author}{Wiggers, K.L.}, \bibinfo{author}{Britto~Jr., A.S.},
  \bibinfo{author}{Koerich, A.L.}, \bibinfo{author}{Heutte, L.},
  \bibinfo{author}{Oliveira, L.E.S.}, \bibinfo{year}{2019}.
\newblock \bibinfo{title}{Image retrieval and pattern spotting using siamese
  neural network}, in: \bibinfo{booktitle}{Intl Joint Conf Neural Networks},
  \bibinfo{address}{Budapest}. pp. \bibinfo{pages}{1--10}.
\bibitem[{Wu et~al.(2017a)Wu, Xu, Zhang, Yan and Ma}]{8302003}
\bibinfo{author}{Wu, H.}, \bibinfo{author}{Xu, Z.}, \bibinfo{author}{Zhang,
  J.}, \bibinfo{author}{Yan, W.}, \bibinfo{author}{Ma, X.},
  \bibinfo{year}{2017}a.
\newblock \bibinfo{title}{Face recognition based on convolution siamese
  networks}, in: \bibinfo{booktitle}{Intl Congress Image Sig Proc, BioMedical
  Eng Inform}, pp. \bibinfo{pages}{1--5}.
\bibitem[{Wu et~al.(2017b)Wu, Oerlemans, Bakker and Lew}]{Wu2017}
\bibinfo{author}{Wu, S.}, \bibinfo{author}{Oerlemans, A.},
  \bibinfo{author}{Bakker, E.M.}, \bibinfo{author}{Lew, M.S.},
  \bibinfo{year}{2017}b.
\newblock \bibinfo{title}{Deep binary codes for large scale image retrieval}.
\newblock \bibinfo{journal}{Neurocomputing} .
\bibitem[{Xu et~al.(2017)Xu, Shen, Xu, Gao, Wang and Tan}]{Xu201745}
\bibinfo{author}{Xu, Y.}, \bibinfo{author}{Shen, F.}, \bibinfo{author}{Xu, X.},
  \bibinfo{author}{Gao, L.}, \bibinfo{author}{Wang, Y.}, \bibinfo{author}{Tan,
  X.}, \bibinfo{year}{2017}.
\newblock \bibinfo{title}{Large-scale image retrieval with supervised sparse
  hashing}.
\newblock \bibinfo{journal}{Neurocomputing} \bibinfo{volume}{229},
  \bibinfo{pages}{45 -- 53}.
\bibitem[{Yue-Hei et~al.(2015)Yue-Hei, Yang and Davis}]{joe}
\bibinfo{author}{Yue-Hei, J.}, \bibinfo{author}{Yang, N.F.},
  \bibinfo{author}{Davis, L.S.}, \bibinfo{year}{2015}.
\newblock \bibinfo{title}{Exploiting local features from deep networks for
  image retrieval}, in: \bibinfo{booktitle}{IEEE Conf Comp Vision Patt Recog},
  pp. \bibinfo{pages}{53--61}.
\bibitem[{Zhu and Doermann(2007)}]{LogoDetection-ICDAR07}
\bibinfo{author}{Zhu, G.}, \bibinfo{author}{Doermann, D.},
  \bibinfo{year}{2007}.
\newblock \bibinfo{title}{Automatic document logo detection}, in:
  \bibinfo{booktitle}{9th Intl Conf Doc Anal Recog}, pp.
  \bibinfo{pages}{864--868}.
\bibitem[{Zhuang et~al.(2015)Zhuang, Cheng, Luo, Pan and He}]{Fuzhen}
\bibinfo{author}{Zhuang, F.}, \bibinfo{author}{Cheng, X.},
  \bibinfo{author}{Luo, P.}, \bibinfo{author}{Pan, S.J.}, \bibinfo{author}{He,
  Q.}, \bibinfo{year}{2015}.
\newblock \bibinfo{title}{Supervised representation learning: Transfer learning
  with deep autoencoders}, in: \bibinfo{booktitle}{24th Intl Joint Conf Art
  Intell}, pp. \bibinfo{pages}{4119--4125}.

\end{thebibliography}

\end{document}